\documentclass{article}
\usepackage{ijcai26}

\usepackage{times}
\usepackage{soul}
\usepackage{url}
\usepackage[hidelinks]{hyperref}
\usepackage[utf8]{inputenc}
\usepackage[small]{caption}
\usepackage{graphicx}
\usepackage{amsmath}
\usepackage{amsthm}
\usepackage{booktabs}
\usepackage{algorithm}
\usepackage{algorithmic}
\usepackage[switch]{lineno}

\usepackage{natbib}

\usepackage{latexsym}

\title{Neuro-Symbolic Agents for Regulated Process Automation: \\ Challenges and Research Agenda}

\author{
Alexander Rombach$^1$\And
Chantale Lauer$^{1,2}$\And
Nijat Mehdiyev$^1$
\affiliations
$^1$ German Research Center for Artificial Intelligence (DFKI), Campus D3 2, 66123 Saarbrücken, Germany\\
$^2$ Saarland University, Campus D3 2, 66123 Saarbrücken, Germany\\
\emails
\{alexander\_michael.rombach, chantale.lauer, nijat.mehdiyev\}@dfki.de
}

\begin{document}

\maketitle

\begin{abstract}
    LLM-based agents are entering regulated industries where they automate judgment intensive quality management processes. We argue that symbolic structures already embedded in these domains, including regulations, typed process models, and compliance constraints, should be treated not merely as external monitoring mechanisms but as core architectural components that shape the agent’s decision-making and behavior. We propose compliance-by-construction as a complementary paradigm to guardrail-based monitoring: a structural foundation that prevents control-flow violations, while guardrails remain essential for catching semantic errors. We identify a structured set of neuro-symbolic research challenges on foundational and capability level and show that addressing them jointly enables compliance-by-construction. We call on the neuro-symbolic community to engage with regulated process automation as a high impact research domain.
\end{abstract}

\section{Introduction}

Neuro-symbolic AI has matured into a productive research paradigm. \citet{garcez2023neurosymbolic} position it as the ``third wave'' of AI, integrating robust neural learning with symbolic reasoning and explainability. Surveys document progress across planning, knowledge representation, and natural-language understanding \citep{hitzler2022neurosymbolic,marra2024statistical,wan2024cognitive}, while a systematic review of 167 publications from 2020--2024 identifies learning/inference (63\%) and knowledge representation (44\%) as dominant themes, with explainability (28\%) and metacognition (5\%) significantly underexplored \citep{colelough2025neurosymbolic}. Yet a persistent concern is the scarcity of applied domains where neuro-symbolic integration is not only beneficial but rather \emph{necessary}, particularly where the stakes of failure are high enough to justify the additional architectural complexity.

We argue that regulated process automation is such a domain. In industries like pharmaceutical manufacturing, biotechnology, and medical devices, quality management (QM) processes are governed by dense regulatory frameworks, such as EU-GMP, ISO 13485, and the EU AI Act (Regulation (EU) 2024/1689), and executed through formally modeled workflows in electronic quality management systems (eQMS). These processes combine a rich symbolic scaffold of regulations, typed process models, and compliance constraints with activities that require contextual, evidence-based reasoning over unstructured documents, which LLMs excel at.

The standard approach to deploying LLM agents in such environments is to add guardrail schema validators that check agent outputs, governance monitors that detect drift, and human-in-the-loop (HITL) gates at critical decision points \citep{derouiche2025agentic}. Recent work on symbolic guardrails demonstrates their value: ShieldAgent \citep{chen2025shieldagent} extracts safety policies into probabilistic logic rule circuits for trajectory verification, while QuadSentinel \citep{chen2025quadsentinel} addresses multi-agent safety via sequent-based supervisory control. However, they address compliance in a post-hoc manner: they monitor and catch violations but cannot structurally prevent them. In high-risk regulated processes where a detected-but-occurred violation is itself a regulatory finding, an additional architectural layer is needed to reduce the range of failures that guardrails must catch.

\paragraph{Our thesis.} The symbolic structures already present in regulated industries should not serve only as an external monitoring layer but as \emph{primary architectural components} that co-constitute the agent's reasoning. We argue for compliance-by-construction as a complementary paradigm: an architectural foundation that structurally prevents control-flow violations (wrong sequencing, missing approvals, skipped mandatory steps), while guardrails remain essential for catching semantic errors that no structural guarantee can eliminate. Realizing this layered approach surfaces a structured set of neuro-symbolic research challenges that we identify in this paper.

Our discussion draws on the consortium research project that develops multi-agent AI for pharmaceutical quality management. While we draw domain context, the challenges we identify are relevant to any scenario where autonomous agents must operate under formal regulatory constraints.

\section{Why Regulated Process Automation Structurally Demands Neuro-Symbolic Integration}

\subsection{Domain Context: Pharmaceutical Quality Management}

Consider deviation management, a core QM process governed by EU-GMP Annex 15 and ISO 13485 \S8.5.2. When a nonconformity is detected in pharmaceutical manufacturing (e.g., a raw-material contamination during an incoming goods inspection, an out-of-specification test result, an equipment malfunction during batch production), a structured workflow is initiated. The deviation is captured and classified by severity; immediate containment actions are defined to prevent further impact; root-cause analysis (RCA) is performed using structured methods such as Ishikawa diagrams, 5-Whys, or 8D reports; corrective and preventive actions (CAPA) are proposed, approved by designated authorities, implemented, and verified for effectiveness; and the case is formally closed with a complete, tamper-proof audit trail.

This process is conceptually modeled in Business Process Model and Notation (BPMN) and executed through eQMS platforms. It involves multiple roles (quality manager, subject matter expert, CAPA owner, approver), multiple artifact types (deviation records, investigation reports, CAPA plans, effectiveness reviews), and multiple decision points where regulatory constraints must be satisfied. Critically, separation-of-duties requirements mandate that the person who investigates a deviation cannot be the same person who approves the resulting CAPA.

The economic stakes are substantial. Industry estimates place the cost of poor quality in the pharmaceutical sector at 25--40\% of operating expenses \citep{bsi2021costquality,dean2004compliance}, significantly exceeding comparable quality-driven industries such as semiconductors (4--8\%). The FDA completed 2,953 CGMP drug inspections worldwide in FY2023, with warning letter frequency rising 43\% per 100 inspections between 2019--2023 \citep{acg2025costquality}. A single compliance failure can trigger warning letters, consent decrees, or product recalls with direct patient-safety consequences.

\subsection{Three Co-Present Properties}

What makes this domain structurally distinctive, and therefore distinctively suited to neuro-symbolic research, is the co-presence of three properties that rarely coincide:

\textbf{Pre-existing symbolic structure.} Regulations, standard operating procedures (SOPs), BPMN models, controlled vocabularies, and typed document schemas constitute a symbolic knowledge base that is already maintained, versioned, and audited within the organization. Unlike many AI domains where symbolic structure must be discovered or learned from data, in this context, it already exists as an organizational artifact. The challenge is to make it computationally active within an agent architecture.

\textbf{Demanding neural tasks.} Each process activity requires reasoning no rule engine can perform: reading a supplier's corrective-action response letter and judging its adequacy against the identified root cause, synthesizing a root-cause hypothesis from a combination of laboratory reports, three prior deviation records involving the same raw material, and an audit finding from the previous quarter, or assessing whether a proposed CAPA addresses the systemic cause rather than the immediate symptom. These require natural-language understanding, multi-document reasoning, and domain judgment, which are precisely the areas where LLMs can provide support.

\textbf{Hard verification requirements with legal consequences.} Compliance properties are binary requirements with regulatory force. The EU AI Act, with high-risk obligations taking binding effect on August 2, 2026, adds further layers: Article 9 mandates risk management systems throughout the AI lifecycle, Article 14 requires human oversight mechanisms that are \emph{technically embedded} in the system (not merely procedural), and Article 17 demands documented quality management procedures \citep{euaiact2024}. For AI systems used as safety components in medical devices, third-party conformity assessment is required under Annex II.

By contrast, most LLM agent deployment domains (e.g., web navigation, open-ended task completion, conversational assistance) offer demanding neural tasks but lack preexisting symbolic structure and hard verification requirements. This makes regulated process automation a uniquely demanding testbed for neuro-symbolic integration where the research challenge is not discovering symbolic structure but \emph{integrating} it with neural capabilities under formal guarantees.

\subsection{Why Classical BPM Is Insufficient}

An objection arises naturally: classical business process management (BPM) solved process compliance decades ago through workflow engines and conformance checking \citep{vanderaalst2016}. What is new?

The answer is that classical BPM assumes \emph{human workers} or \emph{deterministic scripts} at activity nodes. The process engine governs control flow; the executor, whether human or scripted, is either trusted (human judgment, verified post-hoc) or trivially verifiable (script output matches specification). The introduction of LLM agents as activity executors reopens the compliance problem in a new way, because the executor is now probabilistic, opaque, and capable of generating outputs that are syntactically valid but semantically wrong. An LLM might produce a well-formatted severity classification that is factually incorrect, or a grammatically perfect CAPA proposal that fails to address the root cause. The process engine can verify that the activity \emph{occurred} and that its output matches a type schema, but it cannot verify semantic adequacy, nor can a post hoc guardrail do so reliably beyond probabilistic estimation.

This creates challenges that neither BPM alone nor LLM-agent research alone can address. They sit precisely at the neuro-symbolic boundary: the symbolic process structure must constrain the neural executor, the neural executor must provide calibrated signals such that the symbolic layer can reason about its trustworthiness. The interface between them must enforce contracts rich enough to catch meaningful errors but tractable enough to be verified automatically.

\section{Research Challenges}

\begin{figure*}[t]
\centering
\includegraphics[width=\textwidth]{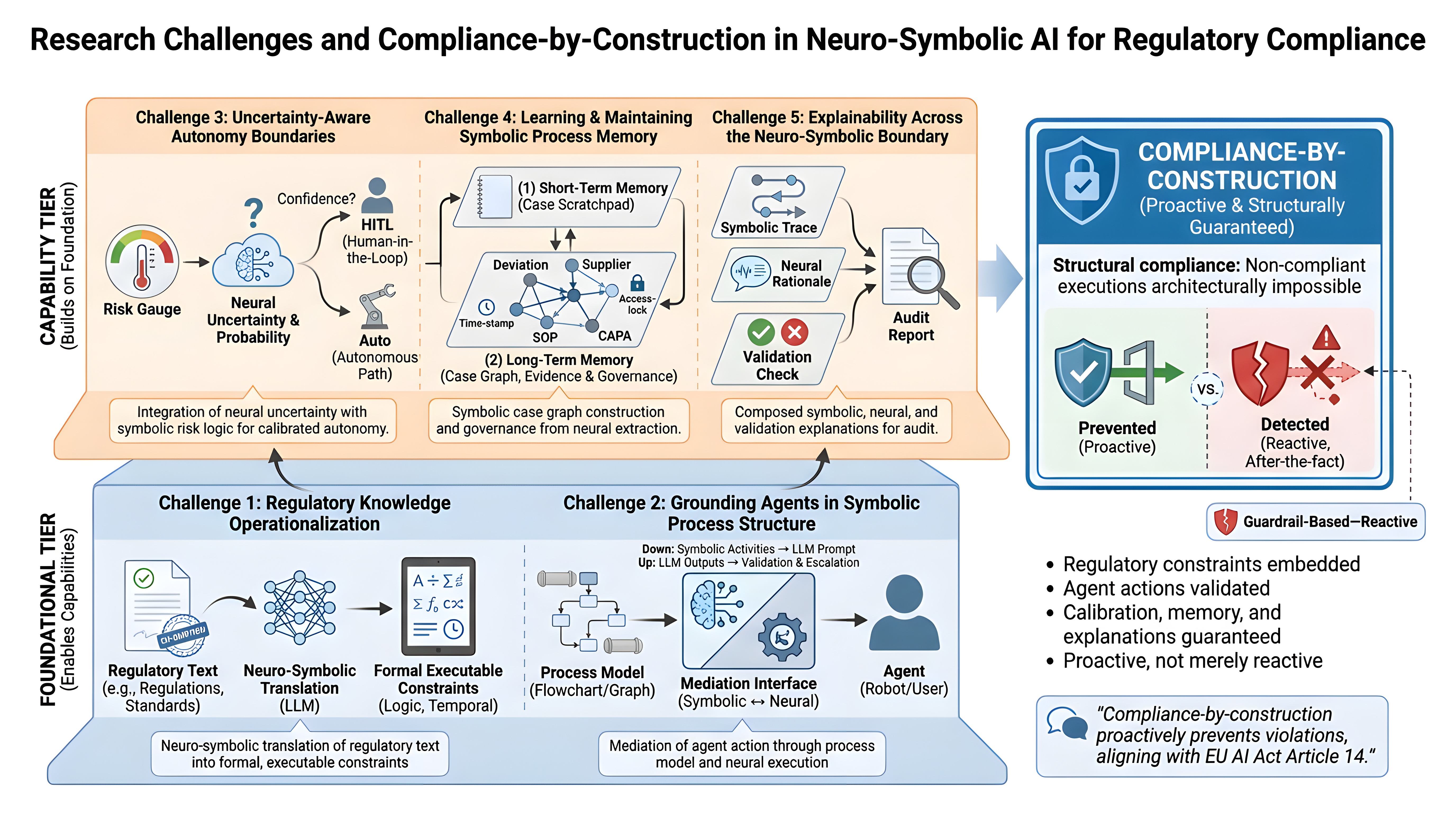}
\caption{Two-tier challenge architecture. The \emph{foundational tier} (Challenge 1: regulatory operationalization; Challenge 2: symbolic process grounding with mediation interface) defines the structural base. The \emph{capability tier} (Challenge 3: uncertainty-aware autonomy; Challenge 4: symbolic process memory; Challenge 5: cross-boundary explainability) extends it. Compliance-by-construction emerges as a property of addressing both tiers jointly by proactively preventing violations rather than detecting them post-hoc.}
\label{fig:architecture}
\end{figure*}

We organize the challenges into two tiers (Figure~\ref{fig:architecture}). The \textbf{foundational} tier (\S\ref{sec:ch1}--\ref{sec:ch2}) defines the architectural base: how regulatory knowledge becomes executable and how agents are bound to process structure. The \textbf{capability} tier (\S\ref{sec:ch3}--\ref{sec:ch5}) builds on this base, addressing calibrated trust, evolving knowledge, and auditable explanations. This hierarchy reflects a dependency structure: the capability-tier challenges presuppose that the foundational tier has been at least partially addressed.

\subsection{Foundational: Regulatory Knowledge Operationalization}
\label{sec:ch1}

\paragraph{The problem.} Regulatory frameworks express requirements in natural language with domain-specific terminology, implicit preconditions, and context-dependent interpretation. EU-GMP Chapter~1 requires that deviations be investigated to determine the root cause. ISO 13485 \S8.5.2 mandates correction to eliminate detected nonconformities. For these to be enforceable at runtime in an agentic system, they must be translated into formal, executable constraints (e.g., temporal properties, state invariants, transition guards) that a process engine can enforce and a model checker can verify.

\paragraph{Why it is neuro-symbolic.} An LLM can read, interpret, and contextualize regulatory text with reasonable accuracy; but producing a formal constraint with precise operational semantics requires a fundamentally different representational commitment. The challenge is compounded in multi-framework regulatory environments where constraints from EU-GMP, ISO standards, and the EU AI Act interact and must be jointly satisfied---a constraint composition problem with no existing solution.

\paragraph{Current state and gaps.} Work on legal requirements translation demonstrates the feasibility of extracting formal representations from regulatory text \citep{singhal2025legal}. ShieldAgent \citep{chen2025shieldagent} demonstrates LLM-assisted policy formalization into probabilistic logic rules for agent monitoring, showing that the neuro-to-symbolic pipeline is viable for safety policies. The broader literature on neuro-symbolic knowledge extraction from text \citep{hitzler2022neurosymbolic,marra2024statistical} provides foundational methods. However, none of these works address the specific challenge of embedding extracted constraints into \emph{executable process models} within \emph{multi-framework regulatory environments} with \emph{formal verifiability} requirements. The gap is not extraction alone but operationalization: producing constraints that are simultaneously faithful to regulatory intent, formally verifiable, and executable at runtime within a process engine.

\paragraph{Research direction.} Hybrid pipelines where LLMs generate candidate formalizations from regulatory text and symbolic methods verify their consistency, completeness, and faithfulness to regulatory intent. The proposed project's approach of translating regulations into machine-checkable policies motivates this direction; the systematic methodology for reliably achieving it remains an open research problem.

\subsection{Foundational: Grounding Agents in Symbolic Process Structure}
\label{sec:ch2}

\paragraph{The problem.} In current agentic architectures (ReAct, AutoGen, CrewAI), process knowledge typically enters the system through the prompt: the agent is told about the workflow in natural language and expected to follow it. \citet{yao2023react} demonstrated the power of interleaving reasoning and acting, but the reasoning is unconstrained, i.e., the LLM decides what to do next based on its interpretation of instructions. In regulated processes, this is structurally insufficient: the process model, not the LLM, must determine the action space.

\paragraph{The neuro-symbolic principle.} The process model should function as a formal structure that \emph{constrains} the agent's action space: at each process state, the set of enabled activities is determined by the model, not by the LLM's prompt interpretation. The LLM's role is to ground each enabled activity by interpreting unstructured case artifacts and producing typed outputs that satisfy the activity's contract. The symbolic layer determines \emph{what} to do next; the neural layer determines \emph{how} to do it.

Neuro-symbolic integration is centered around a layer between neural and symbolic systems where symbolic process state meets neural execution. Consider a concrete example: when the deviation management process enables the activity \textsc{classify\_severity}, the mediation layer constructs a prompt containing the deviation description, relevant SOP excerpts from the process data store, and historical severity distributions retrieved from a Case Graph. It specifies the output schema (\texttt{MINOR, MAJOR, CRITICAL}) and the constraint that \texttt{CRITICAL} triggers mandatory regulatory notification. The LLM produces its classification; the mediation layer validates the output against the type and checks the notification invariant, and either accepts the result (advancing the process state) or triggers re-prompting with a diagnostic that identifies the specific constraint violation. The process model never enters a state that the LLM chose, but rather only states the model prescribed states.

\paragraph{Current state and gaps.} Neuro-symbolic planning research demonstrates the value of binding LLM reasoning to symbolic action specifications. \citet{tantakoun2025llms} survey the landscape of LLMs as planning formalizers, documenting extensive work on LLM+PDDL integration. \citet{capitanelli2024framework} propose a neurosymbolic (NS) robot planning framework using LLMs with PDDL. \citet{kwon2024fast} combine LLM-based goal decomposition with symbolic planners for faster, more accurate task planning. However, these approaches target single-agent, goal-directed plan synthesis. Regulated process automation differs in three respects: processes are long-running rather than one-shot, concurrent and multi-party rather than single-agent, and governed by normative constraints rather than goal states. The extension from plan synthesis to ongoing process execution under regulatory constraints is a significant generalization that the community has not yet undertaken.

\paragraph{Research direction.} Designing the mediation interface: how typed activity specifications are translated into effective prompts, and how LLM outputs are validated against symbolic contracts. The key open problem is the expressiveness-verifiability tradeoff: trivial type checks are verifiable but catch little; rich semantic postconditions (e.g., ``the CAPA addresses the root cause'') are useful but unverifiable by formal methods alone. Identifying the middle ground, perhaps through calibrated confidence scores, lightweight semantic validators, or structured output schemas with partial formal coverage, is a concrete research challenge. We note that the broader perspective of process models as agent world models, with a full formal correspondence between process specifications and world-model components, is a direction we develop separately.

\subsection{Capability: Uncertainty-Aware Autonomy Boundaries}
\label{sec:ch3}

\paragraph{The problem.} Not every process activity warrants the same level of human oversight. Drafting an initial deviation summary from a structured incident report is low-risk and can be fully automated. Confirming a severity classification that triggers mandatory regulatory reporting to national competent authorities is high-risk and demands human approval. Between these extremes lies a continuum in which the appropriate autonomy level depends on case-specific factors: the novelty of the deviation type, the completeness and quality of the available evidence, and the confidence in the LLM's analysis. The EU AI Act's Article~14 explicitly requires human oversight mechanisms ``proportionate to the risks'', which makes calibrated autonomy a legal requirement \citep{euaiact2024}.

\paragraph{Why it is neuro-symbolic.} The risk framework that defines HITL gates and autonomy tiers (e.g., classifying activities by regulatory risk) constitutes symbolic knowledge. The confidence estimate is a neural signal. Adaptive autonomy requires the symbolic layer to reason about neural uncertainty: if the LLM's severity classification carries high uncertainty (a deviation type underrepresented in its experience), the symbolic layer should escalate to HITL regardless of the default risk tier.

\paragraph{Current state and gaps.} Conformal prediction (CP) has emerged as a principled, distribution-free framework for uncertainty quantification in LLM systems \citep{angelopoulos2021gentle}. \citet{ren2023robots} apply CP to calibrate LLM planner confidence for robot task planning, triggering human assistance when prediction sets are non-singleton, thereby  directly demonstrating the viability of UQ-driven autonomy control for LLM agents. \citet{quach2024conformal} extend CP to open-ended language generation. In the process domain specifically, prior work has developed uncertainty-aware predictive process monitoring using CP as well as calibrated execution of enterprise tasks such as business document processing \citep{mehdiyev2023xai,mehdiyev2025augmenting,mehdiyev2025integrating,mehdiyev2025quantifying,majlatow2025uncertainty,Rombach2026}. This establishes methodological foundations for calibrated uncertainty in process-aware systems.

The gap lies in integration: extending these methods from offline process prediction to real-time agent autonomy control. Here, calibrated uncertainty estimates feed into symbolic risk-classification logic that governs HITL escalation during live process execution. The goal is not necessarily that the agent is always correct, but that its autonomy is always calibrated to its demonstrated competence.

\paragraph{Research direction.} Integrating calibrated neural uncertainty (via CP, ensemble disagreement, or epistemic uncertainty decomposition) with symbolic autonomy-control logic in real-time process execution. Such integration allows for the verification of learning-based agents.

\subsection{Capability: Learning and Maintaining Symbolic Process Memory}
\label{sec:ch4}

\paragraph{The problem.} Over time, an organization accumulates a rich history of deviation cases: which raw materials were involved, which suppliers, what root causes were identified, which corrective actions proved effective, which ones failed and had to be revised. This history is invaluable for handling new cases and tasks such as detecting recurring patterns, identifying systemic supplier issues, and calibrating severity assessments against precedent. However, this knowledge exists primarily in unstructured form: investigation reports written in natural language, email threads with suppliers, free-text fields in deviation records.

\paragraph{Why it is neuro-symbolic.} The proposed project models this as a dual-memory architecture: a Short-Term Memory (STM) serving as a case-specific scratchpad for current hypotheses and intermediate evidence, and a Long-Term Memory (LTM) structured as a Case Graph, which represents a knowledge graph (KG) with typed entities (deviation, batch, supplier, SOP, CAPA) connected by domain relations (\emph{caused\_by}, \emph{prevented\_by}, \emph{linked\_to}). Building and maintaining this graph is a paradigmatic instance of learning symbolic abstractions from sub-symbolic data.

The raw inputs are unstructured. Extracting typed entities and relations requires neural processing (named entity recognition, relation extraction, coreference resolution). The resulting graph must be symbolically consistent, meaning that entities must conform to the domain ontology, relations must satisfy type constraints, and the graph must support formal queries that agents rely on at runtime. Moreover, regulated environments impose governance requirements absent from typical KG construction: provenance tracking (which source document did this entity come from?), temporal validity (is this SOP reference still current or has it been superseded?), and access control (which agents can modify which portions of the graph?).

\paragraph{Current state and gaps.} Neuro-symbolic KG construction where neural extraction is guided by symbolic ontologies has shown promise \citep{hitzler2024neurosymbolic}. However, the requirements of regulated environments (e.g., provenance, governance, and the feedback loop where agents query the graph they help build) are largely unaddressed. The Case Graph is not static, but rather an active resource where the root-cause agent searches for precedents, the CAPA agent retrieves prior actions and outcomes, and each agent's outputs feed back in, creating a neuro-symbolic co-learning dynamic.

\paragraph{Research direction.} Neuro-symbolic architectures for KG lifecycle management that jointly satisfy extraction accuracy (neural), ontological consistency (symbolic), provenance (regulatory), and runtime queryability (agentic). The feedback loop between agents and the Case Graph is a compelling instance of neuro-symbolic co-learning.

\subsection{Capability: Explainability Across the Neuro-Symbolic Boundary}
\label{sec:ch5}

\paragraph{The problem.} When a process decision is audited, the explanation must span both layers. The auditor needs to know: (a)~\emph{Why was this activity executed at this point?}---a symbolic question answered by the process model's state and transition logic; (b)~\emph{What evidence did the agent consider and how did it reach its judgment?}---a neural question answered by the LLM's reasoning trace and evidence attributions; and (c)~\emph{Were the formal constraints satisfied?}---a neuro-symbolic question answered by the mediation layer's validation record. No single explainability technique covers all three.

\paragraph{The specific technical challenge.} A symbolic process trace is faithful by construction. An LLM rationale is a post-hoc reconstruction that may not faithfully represent the model's internal computation \citep{turpin2024language}. A cross-boundary framework must reconcile these fidelity levels by indicating which parts are structurally guaranteed and which are best-effort neural attributions. \citet{colelough2025neurosymbolic} find explainability in only 28\% of NS publications. Prior work on XAI for process mining \citep{mehdiyev2021explainable,mehdiyev2024counterfactual} provides a starting point, but extension to real-time multi-layer agent explanations with differentiated fidelity guarantees is open.

\paragraph{Research direction.} Layered explanation architectures that compose symbolic process traces, neural evidence attributions, and contract validation records into audit-ready artifacts with explicit fidelity annotations. Such frameworks would directly address the EU AI Act's transparency requirements (Article 13) for high-risk systems.

\section{Compliance-by-Construction}

Addressing the foundational and capability challenges jointly yields a structural property we call \emph{compliance-by-construction}: an architectural foundation that, by design, makes structural control-flow violations (wrong sequencing, missing approvals, skipped mandatory steps) impossible.

If regulatory requirements are operationalized as process-model invariants (\S\ref{sec:ch1}), and the agent's action space is structurally bound to the process state with validated typed contracts (\S\ref{sec:ch2}), every transition is sanctioned by the model. Calibrated autonomy (\S\ref{sec:ch3}) ensures human oversight at the right moments; governed process memory (\S\ref{sec:ch4}) ensures trustworthy knowledge; cross-boundary explainability (\S\ref{sec:ch5}) ensures auditability.

\paragraph{Complementarity with guardrails.} Compliance-by-construction and guardrails address different failure modes and are complementary. The structural layer prevents deterministic, fully preventable control-flow violations through architectural means. Guardrails (including sophisticated symbolic approaches like ShieldAgent \citep{chen2025shieldagent} and QuadSentinel \citep{chen2025quadsentinel})remain essential for catching semantic errors: an LLM misjudging severity, a poorly reasoned CAPA, or an unforeseen edge case that falls outside the contract specification. Together, they provide defense in depth: the architecture handles what can be guaranteed, and guardrails handle what can only be monitored. The EU AI Act's requirement that human oversight be designed to ``prevent or minimize the risks'' (Article~14(2)) supports this layered approach \citep{euaiact2024}.

Compliance-by-construction guarantees \emph{structural} compliance, not semantic correctness of individual judgments. But it ensures every judgment is made in the right context, validated against formal constraints, documented with provenance, and when confidence is insufficient, routed to human expertise. Guardrails then provide the additional monitoring layer that catches what structural guarantees cannot.

\section{Timeliness}

Three developments make this agenda urgent. First, \textbf{agentic AI is entering regulated production now}. Projects are being built today, and their architectural decisions (guardrails alone vs.\ guardrails plus structural compliance, prompt-based control vs.\ formal grounding) will determine trustworthiness. 
Second, \textbf{the EU AI Act creates legal demand}: high-risk obligations (Articles 9--17) take binding effect on August 2, 2026 \citep{euaiact2024}, imposing risk management, technically embedded human oversight, and documented quality management on AI systems in pharmaceutical and medical device manufacturing. Neuro-symbolic architectures with formal guarantees are becoming legally necessary. 
Third, \textbf{foundational components exist but are not integrated}: neuro-symbolic planning \citep{tantakoun2025llms}, CP for LLM calibration \citep{ren2023robots,angelopoulos2021gentle}, process mining with conformance checking \citep{vanderaalst2016}, uncertainty quantification for process monitoring \citep{mehdiyev2025quantifying}, regulatory NLP \citep{singhal2025legal}, symbolic guardrails \citep{chen2025shieldagent}, and KG construction \citep{hitzler2024neurosymbolic} are all mature. What is missing is the integrative program connecting them.

\section{Discussion and Limitations}

Our proposal has several limitations. First, it assumes formal process models exist and are maintained as authoritative artifacts. This holds for GMP-regulated manufacturing, but not universally. For processes without pre-existing models, a process discovery step would precede our foundational tier, introducing its own neuro-symbolic challenges \citep{vanderaalst2016}.

Second, the added architectural complexity may not be justified for low-risk processes where guardrail-based approaches suffice. Compliance-by-construction is most valuable where the stakes are highest (e.g., patient safety, regulatory reporting).

Third, the expressiveness-verifiability tradeoff (\S\ref{sec:ch2}) remains genuinely unsolved. The most useful postconditions (``the CAPA addresses the root cause'') resist formal verification because they require semantic judgment. The best we can offer architecturally is ensuring semantic judgments occur in the right structural context and escalate to humans when neural confidence is insufficient.

Finally, our challenge framing is deliberately at the research-agenda level, where we identify \emph{what} needs to be solved, not \emph{how}. This is intentional for a position paper aimed at catalyzing community engagement, but the practical feasibility of compliance-by-construction remains to be demonstrated through future work.

\section{Related Work}

\paragraph{Neuro-symbolic AI foundations.} \citet{garcez2023neurosymbolic} position NS~AI as the third wave, emphasizing the integration of learning and reasoning for trust, safety, and explainability. \citet{hitzler2022neurosymbolic} survey neuro-symbolic approaches with a focus on knowledge representation. \citet{marra2024statistical} provide a comprehensive survey bridging statistical relational and NS AI. Our proposal applies this paradigm to a domain in which its core themes are exercised under real-world constraints.

\textbf{LLM-based agent frameworks and symbolic guardrails.} ReAct \citep{yao2023react}, AutoGen \citep{wu2023autogen}, and CrewAI represent the prevailing LLM-as-orchestrator paradigm, providing flexible tool use and multi-agent coordination but no formal compliance guarantees. ShieldAgent \citep{chen2025shieldagent} extracts safety policies into probabilistic logic rule circuits for action trajectory verification. GuardAgent \citep{xiang2024guardagent} synthesizes guardrail code via LLM-driven planning. QuadSentinel \citep{chen2025quadsentinel} addresses multi-agent safety via sequent-based supervisory control. These provide the monitoring layer that remains essential even under compliance-by-construction. Our contribution is the structural foundation that reduces the surface area guardrails must cover.

\paragraph{Neuro-symbolic planning.} \citet{tantakoun2025llms} survey LLMs as planning formalizers, documenting extensive work on LLM+PDDL integration. \citet{capitanelli2024framework} and \citet{kwon2024fast} demonstrate concrete neurosymbolic planning architectures. These target single-agent, goal-directed plan synthesis; extending the symbolic-grounding principle from planning to long-running, multi-party process execution under normative constraints is the generalization we advocate.

\paragraph{LLMs for BPM and trustworthy process AI.} LLMs have been applied to process modeling, task classification, and conversational process modeling \citep{grohs2024large,lauer2025conversational}. Agentic BPM emphasizes governance \citep{vu2025agentic,kirchdorfer2025discovering}, and AI-assisted process mining enables context-sensitive analysis \citep{bruetzke2025aiassisted}. \citet{pfeiffer2025theory} demonstrate trustworthy LLM-driven process automation. \citet{pery2022trustworthy} identify transparency and conformance as prerequisites for trustworthy process mining. Prior work on XAI and UQ for process monitoring \citep{mehdiyev2021explainable,mehdiyev2023xai,mehdiyev2024counterfactual,mehdiyev2025augmenting,mehdiyev2025integrating,mehdiyev2025quantifying,majlatow2025uncertainty} provides methodological foundations for Challenge~3. These address individual capabilities; we advocate the integrated architecture connecting them.

\section{Conclusion}

We have argued that regulated process automation structurally demands neuro-symbolic integration as an architectural necessity. The thesis that symbolic structures should co-constitute the agent's reasoning infrastructure leads to a structured research agenda: foundational challenges of regulatory operationalization and process grounding, supported by capability challenges of calibrated autonomy, symbolic memory, and cross-boundary explainability. Addressing them jointly yields compliance-by-construction, a property that is both academically significant and, with the EU AI Act entering enforcement, increasingly legally required.
The neuro-symbolic community has built the foundations for exactly this kind of integration. Regulated process automation is a domain where those foundations are urgently needed and directly consequential. We invite the community to engage.






\bibliographystyle{named}
\bibliography{references}

@article{garcez2023neurosymbolic,
  title={Neurosymbolic {AI}: The 3rd Wave},
  author={Garcez, Artur d'Avila and Lamb, Lu{\'\i}s C.},
  journal={Artificial Intelligence Review},
  volume={56},
  number={11},
  pages={12387--12406},
  year={2023},
  publisher={Springer}

}

@article{hitzler2022neurosymbolic,
  title={Neuro-symbolic Approaches in Artificial Intelligence},
  author={Hitzler, Pascal and Eberhart, Aaron and Ebrahimi, Monireh and Sarker, Md K. and Zhou, Lu},
  journal={National Science Review},
  volume={9},
  number={6},
  pages={nwac035},
  year={2022},
  publisher={Oxford University Press}
}

@inproceedings{marra2024statistical,
  title     = {From Statistical Relational to Neuro-Symbolic Artificial Intelligence},
  author    = {Raedt, Luc de and Dumančić, Sebastijan and Manhaeve, Robin and Marra, Giuseppe},
  booktitle = {Proceedings of the Twenty-Ninth International Joint Conference on
               Artificial Intelligence, {IJCAI-20}},
  publisher = {International Joint Conferences on Artificial Intelligence Organization},
  editor    = {Christian Bessiere},
  pages     = {4943--4950},
  year      = {2020},
  month     = {7},
  note      = {Survey track},
  doi       = {10.24963/ijcai.2020/688},
  url       = {https://doi.org/10.24963/ijcai.2020/688},
}

@article{wan2024cognitive,
  title={Towards Cognitive {AI} Systems: A Survey and Prospective on Neuro-Symbolic {AI}},
  author={Wan, Zishen and Liu, Che-Kai and Yang, Hanchen and Li, Chaojian and You, Haoran and Fu, Yonggan and Wan, Cheng and Krishna, Tushar and Lin, Yingyan and Raychowdhury, Arijit},
  journal={arXiv preprint arXiv:2401.01040},
  year={2024}
}

@article{colelough2025neurosymbolic,
  title={Neuro-Symbolic {AI} in 2024: A Systematic Review},
  author={Colelough, Brandon C. and Regli, William},
  journal={arXiv preprint arXiv:2501.05435},
  year={2025}
}

@article{chen2025shieldagent,
  title={{ShieldAgent}: Shielding Agents via Verifiable Safety Policy Reasoning},
  author={Zhaorun Chen and Mintong Kang and Bo Li},
  journal={arXiv preprint arXiv:2503.22738},
  year={2025}
}

@article{chen2025quadsentinel,
  title={{QuadSentinel}: Sequent Safety for Machine-Checkable Control in Multi-agent Systems},
  author={Yiliu Yang and Yilei Jiang and Qunzhong Wang and Yingshui Tan and Xiaoyong Zhu and Sherman S. M. Chow and Bo Zheng and Xiangyu Yue},
      year={2025},
      journal={arXiv preprint arXiv:2512.16279},
}

@misc{bsi2021costquality,
  title={Cost of Quality in the Pharmaceutical Sector},
  author={{BSI Group}},
  year={2021},
  howpublished={BSI White Paper}
}

@article{dean2004compliance,
  title={Managing the Cost of Compliance in Pharmaceutical Manufacturing Operations},
  author={Bruttin, Frances and Dean, Doug},
  journal={Pharmaceutical Engineering},
  volume={24},
  number={6},
  year={2004}
}

@misc{acg2025costquality,
  title={Calculating the True Cost of Quality: A {GMP} Perspective for Manufacturing Professionals},
  author={{Auria Consulting Group}},
  year={2025},
  howpublished={ACG Technical Report}
}

@misc{euaiact2024,
  title={Regulation ({EU}) 2024/1689 Laying Down Harmonised Rules on Artificial Intelligence (Artificial Intelligence Act)},
  author={{European Union}},
  year={2024},
  howpublished={Official Journal of the European Union}
}

@book{vanderaalst2016,
  author    = {van der Aalst, Wil M. P.},
  title     = {Process Mining: Data Science in Action},
  edition   = {2},
  year      = {2016},
  publisher = {Springer},
  address   = {Berlin, Heidelberg},
  series    = {Springer Texts in Business and Economics},
  doi       = {10.1007/978-3-662-49851-4},
  isbn      = {978-3-662-49850-7},
  url       = {https://doi.org/10.1007/978-3-662-49851-4}
}

@INPROCEEDINGS{singhal2025legal,
  author={Singhal, Anmol and Breaux, Travis},
  booktitle={2025 IEEE 33rd International Requirements Engineering Conference (RE)}, 
  title={Legal Requirements Translation from Law}, 
  year={2025},
  volume={},
  number={},
  pages={205-217},
  doi={10.1109/RE63999.2025.00028}}

@inproceedings{yao2023react,
  title={{ReAct}: Synergizing Reasoning and Acting in Language Models},
  author={Yao, Shunyu and Zhao, Jeffrey and Yu, Dian and Du, Nan and Shafran, Izhak and Narasimhan, Karthik and Cao, Yuan},
  booktitle={Proceedings of the International Conference on Learning Representations (ICLR)},
  year={2023}
}

@inproceedings{tantakoun2025llms,
  title={{LLMs} as Planning Formalizers: A Survey for Leveraging Large Language Models to Construct Automated Planning Models},
  author={Tantakoun, Marcus and Muise, Christian and Zhu, Xiaodan},
  booktitle={Findings of the Association for Computational Linguistics: ACL 2025},
  pages={25167--25188},
  year={2025}
}

@article{capitanelli2024framework,
  title={A Framework for Neurosymbolic Robot Action Planning using Large Language Models},
  author={Capitanelli, Alessio and Mastrogiovanni, Fulvio},
  journal={Frontiers in Neurorobotics},
  volume={18},
  pages={1342786},
  year={2024}
}

@article{kwon2024fast,
  title={Fast and Accurate Task Planning using Neuro-Symbolic Language Models and Multi-level Goal Decomposition},
  author={Kwon, Minseo and Kim, Yaesol and Kim, Young J.},
  journal={arXiv preprint arXiv:2409.19250},
  year={2025}
}

@article{angelopoulos2021gentle,
  title={A Gentle Introduction to Conformal Prediction and Distribution-Free Uncertainty Quantification},
  author={Angelopoulos, Anastasios N. and Bates, Stephen},
  journal={arXiv preprint arXiv:2107.07511},
  year={2021}
}

@inproceedings{ren2023robots,
  title={Robots That Ask for Help: Uncertainty Alignment for Large Language Model Planners},
  author={Allen Z. Ren and Anushri Dixit and Alexandra Bodrova and Sumeet Singh and Stephen Tu and Noah Brown and Peng Xu and Leila Takayama and Fei Xia and Jake Varley and Zhenjia Xu and Dorsa Sadigh and Andy Zeng and Anirudha Majumdar},
  booktitle={Proceedings of the Conference on Robot Learning (CoRL)},
  year={2023}
}

@inproceedings{quach2024conformal,
  title={Conformal Language Modeling},
  author={Quach, Victor and Fisch, Adam and Schuster, Tal and Yala, Adam and Sohn, Jae H. and Jaakkola, Tommi S. and Barzilay, Regina},
  booktitle={Proceedings of the International Conference on Learning Representations (ICLR)},
  year={2024}
}

@inproceedings{mehdiyev2023xai,
  title={Explainable Artificial Intelligence Meets Uncertainty Quantification for Predictive Process Monitoring},
  author={Mehdiyev, Nijat and Majlatow, Maxim and Fettke, Peter},
  booktitle={PMAI 2023 Workshop at IJCAI},
  pages={29--32},
  year={2023},
  publisher={CEUR-WS}
}

@article{mehdiyev2025augmenting,
  title={Augmenting Post-hoc Explanations with Uncertainty Quantification via Conformalized {Monte Carlo} Dropout},
  author={Mehdiyev, Nijat and Majlatow, Maxim and Fettke, Peter},
  journal={Data \& Knowledge Engineering},
  volume={156},
  pages={102402},
  year={2025}
}

@article{mehdiyev2025integrating,
  title={Integrating Permutation Feature Importance with Conformal Prediction for Robust {XAI}},
  author={Mehdiyev, Nijat and Majlatow, Maxim and Fettke, Peter},
  journal={Engineering Applications of Artificial Intelligence},
  volume={149},
  pages={110363},
  year={2025}
}

@article{mehdiyev2025quantifying,
  author    = {Mehdiyev, Nijat and Majlatow, Maxim and Fettke, Peter},
  title     = {Quantifying and Explaining Machine Learning Uncertainty in Predictive Process Monitoring: An Operations Research Perspective},
  journal   = {Annals of Operations Research},
  year      = {2024},
  volume    = {347},
  number    = {2},
  pages     = {991--1030},
  doi       = {10.1007/s10479-024-05943-4},
  url       = {https://doi.org/10.1007/s10479-024-05943-4},
  issn      = {1572-9338},
  publisher = {Springer US}
}

@article{majlatow2025uncertainty,
  title={Uncertainty-Aware Predictive Process Monitoring in Healthcare},
  author={Majlatow, Maxim and Shakil, Faraz A. and Emrich, Andreas and Mehdiyev, Nijat},
  journal={Applied Sciences},
  volume={15},
  number={14},
  year={2025}
}

@incollection{mehdiyev2021explainable,
  author = {Mehdiyev, Nijat and Fettke, Peter},
  editor = {Pedrycz, Witold and Chen, Shyi-Ming},
    title = {Explainable Artificial Intelligence for Process Mining: A General Overview and Application of a Novel Local Explanation Approach for Predictive Process Monitoring},
    booktitle = {Interpretable Artificial Intelligence: A Perspective of Granular Computing},
    edition = {1st ed. 2021},
    year = {2021},
    volume = {937},
    chapter = {1},
    pages = {1--28},
    publisher = {Springer},
    isbn = {9783030649487}
}

@article{mehdiyev2024counterfactual,
  title={Counterfactual Explanations in the Big Picture},
  author={Mehdiyev, Nijat and Majlatow, Maxim and Fettke, Peter},
  journal={Cognitive Computation},
  volume={16},
  number={5},
  pages={2674--2700},
  year={2024}
}

@article{hitzler2024neurosymbolic,
  title={Neuro-symbolic {AI} and the Semantic Web},
  author={Hitzler, Pascal and Ebrahimi, Monireh and Sarker, Md K. and Daria Stepanova},
  journal={Semantic Web},
  volume={11},
  number={1-3},
  year={2024}
}

@inproceedings{turpin2024language,
  title={Language Models Don't Always Say What They Think: Unfaithful Explanations in Chain-of-Thought Prompting},
  author={Turpin, Miles and Michael, Julian and Perez, Ethan and Bowman, Samuel R.},
  booktitle={Advances in Neural Information Processing Systems (NeurIPS)},
  year={2023}
}

@article{wu2023autogen,
  title={{AutoGen}: Enabling Next-Gen {LLM} Applications via Multi-Agent Conversation},
  author={Wu, Qingyun and Bansal, Gagan and Zhang, Jieyu and Wu, Yiran and Li, Beibin and Zhu, Erkang and Jiang, Li and Zhang, Xiaoyun and Zhang, Shaokun and Liu, Jiale and Awadallah, Ahmed H. and White, Ryen W. and Burger Doug and Wang, Chi},
  journal={arXiv preprint arXiv:2308.08155},
  year={2023}
}

@article{xiang2024guardagent,
  title={{GuardAgent}: Safeguard {LLM} Agents by a Guard Agent via Knowledge-Enabled Reasoning},
  author={Xiang, Henry and Zheng, Linzhi and Hing, Junyuan and Li, Qinbin and Xie, Han and Zhang, Jiawei and Xiong, Zidi and Xie, Chulin and Yang, Carl and Song, Dawn anf Li, Bo},
  journal={arXiv preprint arXiv:2406.09187},
  year={2024}
}

@inproceedings{grohs2024large,
  title={Large Language Models Can Accomplish Business Process Management Tasks},
  author={Grohs, Michael and Abb, Lukas and Elsayed, Nourhan and Rehse, Jana-Rebecca},
  booktitle={BPM Workshops},
  pages={453--465},
  year={2024},
  publisher={Springer}
}

@incollection{lauer2025conversational,
author = "Lauer, Chantale and Pfeiffer, Peter and Rombach, Alexander and Mehdiyev, Nijat",
title = "Conversational Business Process Modeling using LLMs: Initial Results and Challenges",
year = 2025,
doi = "10.18420/EMISA2025_03",
booktitle = "EMISA 2025",
publisher = "Gesellschaft für Informatik e.V.",
address = "Bonn",
pissn = "P-362",
pages = "P3+2--O4+5",
}

@inproceedings{vu2025agentic,
  title={Agentic {BPM}: Practitioner Perspectives on Agent Governance},
  author={Vu, Hoang and Klievtsova, Nataliia and Leopold, Henrik and Rinderle-Ma, Stefanie and Kampik, Timotheus},
  booktitle={BPM: Responsible BPM Forum},
  pages={29--43},
  year={2025},
  publisher={Springer}
}

@article{kirchdorfer2025discovering,
  title={Discovering Multi-agent Systems for Resource-centric Business Process Simulation},
  author={Kirchdorfer, Lukas and Bl{\"u}mel, Robert and Kampik, Timotheus and van der Aa, Han and Stuckenschmidt, Heiner},
  journal={Process Science},
  volume={2},
  number={1},
  pages={4},
  year={2025}
}

@inproceedings{pfeiffer2025theory,
  title={From Theory to Practice: Real-World Use Cases on Trustworthy {LLM}-Driven Process Modeling, Prediction and Automation},
  author={Pfeiffer, Peter and Rombach, Alexander and Majlatow, Maxim and Mehdiyev, Nijat},
  booktitle={Proceedings of the ACM SIGMOD International Conference on Management of Data},
  year={2025}
}

@inproceedings{pery2022trustworthy,
  title={Trustworthy {AI} and Process Mining: Challenges and Opportunities},
  author={Pery, Andrew and Rafiei, Majid and Simon, Michael and van der Aalst, Wil M. P.},
  booktitle={Process Mining Workshops},
  pages={395--407},
  year={2022},
  publisher={Springer}
}

@article{derouiche2025agentic,
  title={Agentic {AI} Frameworks: Architectures, Protocols, and Design Challenges},
  author={Derouiche, Hana and Brahmi, Zaki and Mazeni, Haithem},
  journal={arXiv preprint arXiv:2508.10146},
  year={2025}
}

@misc{bruetzke2025aiassisted,
  author    = {Br{\"u}tzke, Philipp and Killewald, Robin and Franzoi, Stefano and vom Brocke, Jan},
  title     = {AI-Assisted Process Mining for Context-Sensitive Analysis Support},
  booktitle = {Proceedings of the 33rd European Conference on Information Systems},
  editor    = {Krasnova, Hanna and Kranz, Johann and Lindgren, Rikard},
  pages     = {1--16},
  year      = {2025},
  publisher = {AIS eLibrary},
  address    = {United States}
}

@article{Rombach2026,
author = {Rombach, Alexander and Mehdiyev, Nijat},
doi = {10.1007/s10032-026-00572-y},
issn = {1433-2833},
journal = {International Journal on Document Analysis and Recognition (IJDAR)},
month = {mar},
title = {{Beyond Accuracy: Understanding Model Confidence in Key Information Extraction with Conformal Prediction}},
url = {https://doi.org/10.1007/s10032-026-00572-y https://link.springer.com/10.1007/s10032-026-00572-y},
year = {2026}
}

\end{document}